# A Deep Q-Learning based Smart Scheduling of EVs for Demand Response in Smart Grids

**Viorica Rozina Chifu, Tudor Cioara\*, Cristina Bianca Pop, Horia Rusu and Ionut Anghel**

Computer Science Department, Technical University of Cluj-Napoca, Memorandumului 28, 400114 Cluj-Napoca, Romania; viorica.chifu@cs.utcluj.ro; cristina.pop@cs.utcluj.ro; rusu.io.horia@student.ut-cluj.ro; ionut.anghel@cs.utcluj.ro;

\* Correspondence: tudor.cioara@cs.utcluj.ro

**Abstract:** Economic and policy factors are driving the continuous increase in the adoption and usage of electrical vehicles (EVs). However, despite being a cleaner alternative to combustion engine vehicles, EVs have negative impacts on the lifespan of microgrid equipment and energy balance due to increased power demand and the timing of their usage. In our view grid management should leverage on EVs scheduling flexibility to support local network balancing through active participation in demand response programs. In this paper, we propose a model-free solution, leveraging Deep Q-Learning to schedule the charging and discharging activities of EVs within a microgrid to align with a target energy profile provided by the distribution system operator. We adapted the Bellman Equation to assess the value of a state based on specific rewards for EV scheduling actions and used a neural network to estimate Q-values for available actions and the epsilon-greedy algorithm to balance exploitation and exploration to meet the target energy profile. The results are promising showing that the proposed solution can effectively schedule the EVs charging and discharging actions to align with the target profile with a Person coefficient of 0.99, handling effective EVs scheduling situations that involve dynamicity given by the e-mobility features, relying only on data with no knowledge of EVs and microgrid dynamics.

**Keywords:** Deep Q-Learning; EVs scheduling; Vehicle to Grid; Demand Response; Reinforcement Learning; Model Free Optimization.

## 1. Introduction

The ongoing transition towards various economic sectors' decarbonization enables fossil energy supply substitution with renewable energy. However, the rapid adoption of small-scale renewable at the edge of the grid makes the smart grid management process more complex and exposed to uncertainty related to renewable production [1, 2]. The digitization and decentralization principles bring to the forefront the energy demand flexibility as a key support to accommodate high shares of variable renewable energy [3, 4]. Leveraging local flexibility is possible to maintain a balance between supply and demand at lower costs using the energy assets of the citizens rather than the ones owned by the grid operator that are more expensive to operate [5-7].

The challenges are even more evident and difficult to tackle in the context of the increased adoption of electrical vehicles (EVs) [8]. In that respect grid management should closely cooperate and interact within a low latency context with EVs coordination and aggregation services to procure their energy scheduling flexibility to support local network balancing or to achieve self-sufficiency [9, 10]. However, EVs usage has several shortcomings such as the limited battery range, relatively short battery lifespan, averaging 10-20 years or up to 150,000 miles, and the lack of existing infrastructure for charging electric vehicles. Other major issues refer to the impact of EVs on the power grid, encompassing factors such as the rise in short-circuit currents, deviations in voltage levels beyond standard limits, and the potential impact on the lifespan of equipment due to increased power demand [11]. Studies have shown that fully charging an electric vehicle uses the same energy needed to power a home during peak energy use times and nowadays the energy grid cannot deal with a major or sudden spike in EVs usage [12]. Moreover, it is not only a problem related to the quantity of electricity consumed by EVs but also the timing of their usage [13]. If many electric vehicles were to be charged in the evening, the grid could experience significant strain, potentially leading to blackouts.

To address the challenges of EVs coordination for smart grid management, solutions are being proposed to convince people to change their behavior [14]. Dynamic traffic for electricity is well discussed in the literature as a solution to move the charging schedules of EVs to moments of the day with extra solar energy [15]. However, it still faces open challenges related to the complexity of the pricing structure that involves hourly or time-of-use rates, making EV owners reluctant to engage in or the fact that requires EV owners to adapt their behavior based on fluctuating electricity prices [16]. Other solutions are leveraging on the setup of Demand Response (DR) programs for EVs aiming to manage and modify their charging or discharging patterns based on grid conditions [17]. However, this presents a major challenge: the need to balance energy production and consumption in the face of intermittent and non-programmable energy sources which are strongly influenced by atmospheric conditions, making it difficult to predict [2]. Many existing state-of-the-art approaches consider the EVs participation in DR as a model-based constraint satisfaction problem [18]. They achieve good results; however, their main drawbacks are related to the model's assumptions and structure, which need to be continuously updated to meet the actual conditions [19]. Moreover, the model solving is computationally complex, being less suitable for large-scale dynamic features involved in EVs scheduling [20].

In this paper, we address the limitations by proposing a model-free solution for EVs participation in DR programs, leveraging Deep Q-Learning. Our objective is to schedule the charging and discharging activities of EVs within a microgrid to align with a target energy profile provided by the distribution system operator for a specific interval. The Deep Q-Learning solution determines optimal EV scheduling actions based only on the current state, which is represented by the state of charge of EVs and the availability of charging stations in the microgrid. We have adapted the Bellman Equation that plays a crucial role in assessing the value of a specific state considering specific rewards for EV scheduling actions and Q-values to evaluate the effectiveness of each action. The methodology typically involves the estimation of Q-values for available EVs scheduling actions and then the selection of the actions with the highest reward. Traditionally, the Q-function is implemented as a look-up table, updated after each transition. However, in dynamic scenarios with a multitude of complex states, rendering a table insufficient, thus we opt for representing the Q-function using a neural network the input variables signify the state, and the output provides the Q-value for every possible action. For action selection, we have employed the epsilon-greedy algorithm to balance the exploitation of the current best action or explore new possibilities for EVs scheduling to meet the target DR energy profile.

The rest of the paper is structured as follows: Section 2 reviews the literature on EVs scheduling and coordination solutions; Section 3 presents our Deep Q-Learning solution for EVs participation in DR programs; Section 4 provides experimental validation, demonstrating the approach's effectiveness in meeting target energy profiles and evaluating the quality of the learned EVs scheduling solutions. Section 5 concludes the paper and presents future work.

## 2. Related work

Reinforcement learning-based approaches (RL) for EV scheduling have been studied in the literature to optimize the charge and discharge decisions in a dynamic and uncertain environment. Wen et al. [21] schedule electric vehicles for charging and discharging using a Deep Q-Learning-based method that considers both the mobility of electric vehicles and the random charging behavior of users. The EV's scheduling is made by considering a distribution model for user charging times, a model for charging demand, a dynamic model for EV state-of-charge, and a model for travel locations. A dynamic reward function, the electricity price, the charging and discharging cost, and the battery degradation cost are employed to optimally schedule EVs' such that the economic operating cost is reduced, and the efficiency of the charging/discharging process is improved. However, the method suffers from a drawback, namely the dynamic reward function doesn't accurately reflect the benefits of charging/discharging operations, when user behavior deviates from the assumptions made in the travel locations model. Lee et.al. [22] also propose a deep RL-based method for EVs charging and discharging in real time. The aim is to minimize the EVs' charging cost and to reduce grid loads during peak hours. To model the usage patterns of charging stations, a parametric density function estimation is used that allows the learning agents to make optimal charging and discharging decisions. The effectiveness of the approach has been demonstrated through simulation, and the results demonstrated that a significant reduction of the energy cost and the grid load is obtained. Wan et. al. [23] combines a representation network with deep RL to schedule the charging and discharging of electric vehicles in a domestic setting such that the EVs' charging/discharging cost is minimized. The representation network (i.e., LTSM network) is used to extract features from historical electricity prices that together with EV's battery SoC, and user-driving pattern (i.e., the arrival and departure times) are given as inputs to the Q-learning network to estimate the optimal action-value function. Like in the previous work, the effectiveness of the approach has also been demonstrated through simulation. In their work, Viziteu et.al. [24] use Deep Q- learning to schedule EV's for charging by considering the battery SoC, the battery capacity, the longest trip distance, the distance between EV' current location and the charging station, the waiting time until the next charging slot is available, and the charging station power. The aim is to manage electric vehicle congestion by scheduling them in advance at charging stations and optimizing drivers' journeys. Like in the case of previous approaches, the approach's effectiveness has been demonstrated through simulations. Cao et. al. [25] address the problem of scheduling electrical vehicles for charging, aiming to reduce carbon emission costs and peak loads in a community. The proposed approach integrates three algorithms: an offline scheduling algorithm that considers EV profiles, the EVs' random arrivals, and power constraints to minimize the carbon emission cost, an online heuristic rolling-based scheduling algorithm that considers EVs' arrival times and SoC to take carbon emission decisions, and an actor-critic RL scheduling algorithm that updates the EV carbon emission actions. The approach has been tested through simulation. Liu et al. [26] schedule the EVs for charging, to minimize the cost of charging the EVs and stabilizing the voltage of the distribution network. The scheduling and voltage control problems are modeled as a Markov decision process that considers the uncertainty introduced by the charging behavior of electric vehicle drivers and uncontrolled load, as well as fluctuations in energy prices and renewable energy production. The deep deterministic policy gradient algorithm is used to return discrete and continuous control actions. The reward function is defined to ensure a balance between the EV charging and voltage stability. Paraskevas et.al. [27] introduce a win–win Strategy based on deep Q-learning (DLR) aiming to maximize both the charging station's profits and drivers' charging demands. The DRL agent's decisions are taken in real-time by considering the uncertainties due to future EV arrivals distribution and electricity prices. Wang et.al. [28] proposes an improved rainbow-based deep RL strategy for the optimal scheduling of charging stations. The charging process involving the matching between EV charging demand and CS equipment resources is modeled as a finite Markov decision process. To schedule EVs' at charging stations under uncertain conditions due to EV arrival and departure times, a DQN-based rainbow algorithm is used. Li et al. [29] propose an approach for the optimal scheduling of electric vehicles for charging/discharging to minimize charging costs and to ensure full charging of electric vehicles. The EV charging/discharging scheduling problem is defined as a constrained Markov decision process where the constraints are full charging of EVs and minimization of charging cost and is solved using a secure deep RL strategy. An electric vehicle aggregator (EVA) is proposed in [30] as the decision-making authority for scheduling electric vehicles for charging. Scheduling is done considering uncertainty related to renewable energy generation and user demand and has as its main objectives, minimizing the cost of EVA and fluctuations in power exchange of the microgrid. The EV charge scheduling problem is modeled as a Markov decision-making process based on deep RL and is solved using a twin delayed deep deterministic policy gradient algorithm. Heendeniya et.al. [31] introduce a DRL technique based on actor-critical architecture for optimal charging of electric vehicles. The aim is to minimize both charging time and expected voltage violations. The proposed technique can learn an optimal policy by considering the voltage magnitude measurements and improve the scalability by imposing partial observability. Shi et.al. [32] propose a RL-based approach for managing a community's electric vehicle fleet that offers ride-hailing services to residents. The aim is to minimize the clients' waiting time, the electricity cost, and vehicles' operational costs. To solve the problem of dispatching the electric vehicle fleet, decentralized learning is combined with centralized decision-making. Decentralized learning allows EVs to share their knowledge and learned models to estimate the state-value function, while centralized decision-making assures EV coordination to avoid scheduling conflicts. Li et. al. [33] propose a strategy for charging EVs aiming at the main goal of EVs charging cost minimization. The strategy uses JANET to extract regular variability in energy price and deep RL to adjust the charging strategy of EVs based on variability in energy price. Ding et.al. [34] also proposes a strategy for charging electric vehicles to increase the profit of distribution system operators while ensuring the correct use of distribution networks to avoid potential voltage problems. A model based on Markov decision-making is defined to describe the uncertainty of EV user behavior, and a learning algorithm based on deterministic political gradients is used to solve the model. Park et.al [35] solve the problem of scheduling EV for charging through a deep learning-based method of multi-agent reinforcement, in which the training stage is centralized, and the execution stage is decentralized. The aim is to minimize the operating costs of charging stations and to ensure the amount of energy specified for charging the EV.

However, even though the model-free reinforcement learning solutions are promising for EVs scheduling and smart grid management, they have some challenges partially addressed in this paper. Reinforcement learning algorithms require many samples, episodes, and action simulations to learn the optimal EVs scheduling policies. The convergence of rewards and loss to desirable values even in our solution was a process that took a considerable number of learning episodes. Moreover, in most cases, achieving a good balance between exploration (i.e., trying new EVs scheduling actions) and exploitation (i.e., leveraging on known actions) is rather challenging. We have addressed it by employing an epsilon greedy strategy and fine-tuning the delicate balance between exploitation and exploration EV scheduling search space. Another challenge in the studied literature is the reward function design for the objectives of the EVs scheduling. We have addressed it by adapting the Bellman Equation to evaluate the EVs scheduling state from an energy perspective and defined a Q-function to determine the EVs scheduling effectiveness for the DR program. Finally, the state and action spaces in EV scheduling can be high-dimensional and complex, making it challenging for the Q-learning network to efficiently learn the optimal policy for DR, thus requiring fine-tuning of the learning and network architecture parameters.

## 3. Materials and Methods

The objective of the EVs scheduling is to be able to participate collectively in demand response programs (DR) used to balance the supply and demand of electricity in the smart grid [36]. We have considered V2G-enabled EVs (Vehicle to Grid) [37] that can respond to grid operator signals by modifying their energy consumption or generation patterns. We formalize a DR program as:

$$DR = < P = EVs, E_{target}(T) > \tag{1}$$

where $E_{target}$ is a goal energy curve provided to the participant EVs. A Q-learning algorithm determines the collective response of EVs.

In Q-learning one popular assumption is that the environment in which the agent learns is described as a Markov decision process. We will use this simplifying assumption which means that the future state of the environment depends only on the previous state and actions of the agent. We formally define the environment in which the agent acts as:

$$E = < S, A, r, \pi > \tag{2}$$

where, $S = \{S^k\}$ is the state space representing the set of possible states that the agent might be in, $A$ is the action space representing the set of all possible actions that the agent can take when it is in each state, $\pi$ is the policy that the agent learns, and $R$ are the rewards given to the agents as results of his actions.

The state space represents all possible mapping of EVs on charging stations over the demand response program interval $T$. We model it as a matrix with the dimensions $NxT$

$$S^k = \begin{bmatrix} s_{11} & s_{12} & ... & s_{1T} \\ s_{21} & s_{22} & ... & s_{2T} \\ ... & ... & ... & ... \\ s_{N1} & s_{N2} & ... & s_{NT} \end{bmatrix} \tag{3}$$

where $N$ is the number of available charging stations and $T$ is the number of timeslots $t$ for DR. An element of the matrix stores the identified of the EV which is assigned to a charging station at specific timeslot or zero otherwise:

$$s_{it} = \begin{cases} 0, & \text{if no EV is assigned to } cs_i \text{ at } t \\ ev_{ID}, & \text{if EV with } ev_{ID} \text{ is assigned to } cs_i \text{ at } t \end{cases} \tag{4}$$

For each environment state the equivalent energy state is determined and stored in a HashMap $H$ in which the set of keys $K$ are the individual timeslots in the DR program interval $T$ and the list of values $V_E$ represent the energy charged or discharged at each charging point:

$$H_{Energy}: K_T \rightarrow List < V_E > \tag{5}$$

The amount of energy charged or discharged in total by all charging stations in the microgrid per a time slot $t$ is determined as:

$$(\forall)\ t \in K_T, E_{EV}(t) = \sum_{v_e \in V_E} List < V_E > \tag{6}$$

The initial state of the agent exploration process is represented by a matrix with zero values, indicating that no EVs are assigned to any charging station within the interval $T$. The agent will explore and learn by performing actions, receiving rewards based on these actions, and updating the Q-values. An action corresponds to assigning an electric vehicle $ev_{ID}$ to a charging station $cs_i$ in a certain timeslot $t$:

$$a = (cs_i, ev_{ID}, t, \nabla) \tag{7}$$

When the agent performs an action, it results in a new state, which is represented as a new matrix containing the new configuration of electric vehicles assigned to charging stations. This iterative process allows the agent to learn the optimal policy for assigning electric vehicles to charging stations while considering the specific constraints and the objective defined for our problem.

In the defined environment we have considered several constraints on the set of actions the agent may consider driven by the physical limitations of the energy infrastructure. The type of action for an EV at a charging station at a time slot can be either a charge or a discharge:

$$C_1 : \nabla = \{C, D\}, a_{ev_{ID}} = C \vee a_{ev_{ID}} = C \tag{8}$$

An EV will be considered for scheduling only if its current state of charge falls within a predefined range based on its capacity, and it is corelated with the types of actions associated:

$$C_2 : SoC_{ev_{ID}}^{MIN} < SoC_{ev_{ID}} \ll SoC_{ev_{ID}}^{MAX} \rightarrow a_{ev_{ID}} = C \tag{9}$$

$$C_3 : SoC_{ev_{ID}}^{MIN} \ll SoC_{ev_{ID}} < SoC_{ev_{ID}}^{MAX} \rightarrow a_{ev_{ID}} = D \tag{10}$$

The electric vehicle will only be eligible for charging is its $SoC$ is significantly less than the maximum capacity and for discharging if the $SoC$ is significantly higher than the minimum capacity.

Two different EVs $ev_{IDi}$ and $ev_{IDj}$ cannot be scheduled at the at the same charging station $cs$ in the same time slot $t$:

$$C_4 : (\forall) \ cs, ev_{IDi}, ev_{IDj} \rightarrow a(cs, ev_{IDi}, t, \nabla) \vee a(cs, ev_{IDi}, t, \nabla) \tag{11}$$

Finally, the amount of energy charged or discharged by EVs at a timeslot $E_{EV}(t)$ must not exceed the target energy provided by the grid operator in the DR program:

$$C_5 : E_{target}(t) \geq E_{EV}(t) \tag{12}$$

The goal of the agent is to find new environmental states by successively simulating the execution of actions such that the distance between the target energy curve and the amount of energy charged and discharge by EVs is minimized over the DR program interval $T$:

$$\min_T (E_{target} - E_{EV}) \tag{13}$$

Moreover, the agent goal is to learn an optimal policy, $\pi : S \longrightarrow A$ that maximizes the cumulative reward obtained by executing a sequence of actions in the defined environment. To find optimal policy, the agent explores the environment, observes the rewards obtained when taking specific actions in each state, and uses this experience to update his knowledge of the best actions taken in different states. The optimal policy has associated an optimal $Q - function$ that represents the maximum return that can be obtained from a state $S^k$, taking an action $a$ and subsequently following the optimal policy:

$$\pi : S \longrightarrow A, \pi = \{(S^k, a) | Q^*(S^k, a) \ is \ optimal, S^k \in S \wedge a \in A\} \tag{14}$$

The optimal action-value function $Q^*(S^k, a)$ is described by the Bellman equation which establishes a link between the optimal action value of one state-action pair and the optimal action value of the next state-action pair.

$$Q^*(S^k, a) \leftarrow (1 - \alpha) * Q(S^k, a) + \alpha * (r(S^k, a) + \gamma * max_{a'} Q(S^{k'}, a') \tag{15}$$

where $Q^*(S^k, a)$ is the new Q-value calculated for the action $a$, $\alpha$ is the learning rate, $r(S^k, a)$ is the immediate reward, $\gamma$ is the discount rate, and $max_{a'} Q(s', a')$ is the maximum Q-value selected from all Q-values obtained by executing all possible actions in the next state $S^{k'}$. In our model we have set the initial Q-value of a pair $(S^k, a)$ is usually initialized with an arbitrary fixed value before learning begins.

The immediate rewards assigned to a state-action pair is:

$$r(S^k, a) = \begin{cases} MAX_{penalty}, \ if \ a \ \neg C_1 \vee \neg C_2 \neg C_3 \vee \neg C_4 \\ 100 * |E_{target}(T) - EV(T)|, \ where \ EV(T) \ is \ linked \ to \ S^k \end{cases} \tag{16}$$

Figure 1 shows an overview of the Q-Learning model defined for EVs coordination for DR participation.

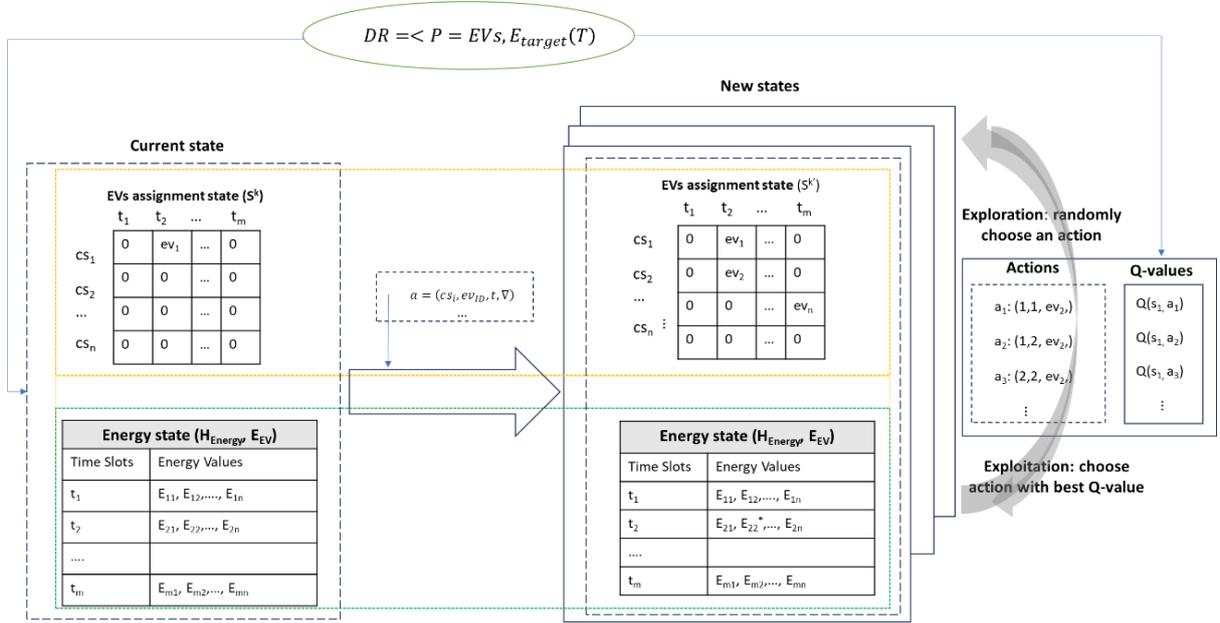

**Figure 1.** The Q-Learning model.

The $Q - function$ is modeled using a deep neural network (Q-network) having as input the states (i.e., EVs mapping to charging stations and time slots states) and outputs the associated Q-value. To train the neural network we use the $\varepsilon$-greedy algorithm [38] to ensure that the exploration-exploitation strategies are balanced. It helps to balance the trade-off between exploring unknown actions and exploiting the ones with the best Q-values. Figure 2 shows the structure of the Q-network used.

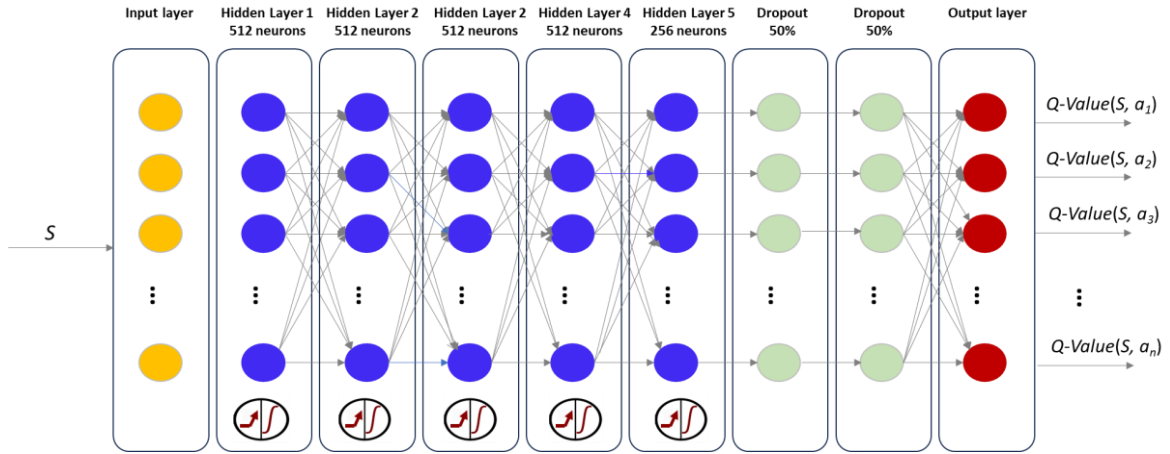

**Figure 2.** Q-network used for Q-function approximation.

The Q-network architecture comprises five hidden dense layers with ReLU activation: four layers with 512 neurons and one with 256 neurons. It also includes two dropout layers, each with a dropout rate of 0.5 to prevent overfitting. There's an input layer matching the size of the state space and an output layer matching the size of the action space, representing the Q-value for each possible action in the environment.

---

**Algorithm 1: Deep Q-learning for EVs scheduling**

**Inputs:** $S$ - state space, $A$ - action space, $CS$ − the of charging stations, $EV$ − the set of electrical vehicles, $T$- the time slots, $N$ - the number of epochs for model training, $E_{target}(T)$ − DR program target curve

**Outputs:** $\theta$ the weights of TQ-network

**Q-network parameters**: $\theta$ − weights of the network, $\varepsilon$ − epsilon parameter, $D$ − replay memory, $m$ - target network update frequency, $\eta$ -learning rate, $\gamma$ − discount factor

**Begin**

1.  $Q − network, TQ − network ← Initialize\ parameters$
2.  $for\ epoch = 1\ to\ N\ do$
3.  $\quad S^{initial} ← generate\ (E_{target}(T), CS, EV)$
4.  $\quad S^{current} = S^{initial}$
5.  $\quad \varepsilon = 1, done = False$
6.  $\quad while\ (\neg done)\ do$
7.  $\quad\quad if\ (random\ (0,1) < \varepsilon)\ do$
8.  $\quad\quad\quad action = select\_random\_action\ (A)$
9.  $\quad\quad else$
10. $\quad\quad\quad action = argmax(Q − network(S^{current}))$
11. $\quad\quad endif$
12. $\quad\quad S^{next} = [done = take\_action(action), reward\ (S^{current}, action)]$
13. $\quad\quad D.append\ (S^{current}, action, reward, S^{next}, done)$
14. $\quad\quad S^{current} = S^{next}$
15. $\quad end\ while$
16. $\quad if\ (|D| > batch\_size)\ then$
17. $\quad\quad mini\_batch = sample\_batch\ (D, batch\_size)$
18. $\quad\quad TQ − values = compute\ (mini\_batch, \gamma, TQ − network)$
19. $\quad\quad Q − values = compute\ (mini\_batch, \gamma, Q − network)$
20. $\quad\quad loss = mean\_squared\_error\ (TQ − values, Q − values)$
21. $\quad\quad \theta = \theta − \eta * \nabla Loss(\theta)$
22. $\quad\quad if\ (epoch\ \%\ m\ == 0)\ then$
23. $\quad\quad\quad TQ − network ← copyWeights(Q − network)$
24. $\quad\quad end\ if$
25. $\quad\quad \varepsilon = \varepsilon * \varepsilon_{decay}$
26. $\quad end\ for$
27. $return\ \theta\ of\ TQ − network$

**End**

---

In the learning process we also have used a target Q-network (TQ-network) with the same architecture as the Q-network. Its weights are updated more slowly and periodically with the weights from the Q-network. The TQ-network is used for stabilizing the learning process and improving the algorithm's efficiency by updating the network after a certain number of iterations instead of after every episode. Both the Q-network architecture and the target Q-network will be initialized with the same randomly generated weights.

The deep Q-learning model for EVs scheduling in demand response is presented in Algorithm 1. The Q-network architecture and the TQ-network are initialized, and the network weights are randomly generated (lines 1-5).

The replay memory is initialized and updated in the learning process with the *tuples* containing the transitions from one state to another, as well as the *rewards* given to these transitions. A Boolean variable is defined for situations in which the training episode is over: either when all electric vehicles have been assigned to a charging station in a specific time slot or when two electric vehicles have been assigned to the same charging station at the same time slot. The batch size variable is used to determine how many entries from the replay memory are randomly selected to be used for training the model.

After completing the initialization, the model training process starts, and the agent interacts with the defined scheduling environment. Initially, it reads the initial state of the environment before the scheduling (line 3-4) which includes the scheduling matrix of electric vehicles at the charging stations, together with the DR program target energy profile. An action is selected based on the $\epsilon$-greedy method to ensure the balance between exploration and exploitation for model training (lines 7-11). A random number between 0 and 1 is generated and then compared to the value of the hyperparameter $\varepsilon$ originally initialized with 1. In case the random value is smaller than $\varepsilon$, the action is chosen randomly, without considering the reward obtained. If the value is greater than $\varepsilon$, the action with the biggest Q-value for the current state is selected. The action selected by the agent is executed by updating the scheduling matrix. The reward is calculated, and the next state is updated with the current EV planning matrix (line 12) and the information is stored in the replay memory. Since the size of the resume memory is fixed, when the number of transitions in memory exceeds its size, the oldest value is removed.

Then for the training process we randomly extract a mini-batch sample of transitions from the replay memory $D$ of the size of the batch (lines 16-24). This step is only done once there are enough transitions in the memory to cover the batch size. The Q-values from TQ-network are determined from stored transitions from the previous step and the Q-values from the Q-network are determined based on the original network (18-19). Afterward, we determine the loss by using the mean squared difference between the TQ-values and Q-values and calculate the loss gradients relative to the Q-network weights (lines 20-21). The new weights of the model are determined considering $\eta$ a hyper-parameter representing the learning rate of the network in which the weights are updated. The weights of the Q-network are updated based on the calculated gradients and after each $m$ episode, the weights of the current Q-network are copied to the target Q-network to improve the accuracy of the target network for future training (lines 22-23). Finally, the value of the $\epsilon$ parameter is updated via a constant decay rate, and the TQ-network parameters are returned.

## 4. Evaluation results

To evaluate and demonstrate the effectiveness and efficiency of our scheduling method, we have used a data set of EVs. For each EV in the data set, the following data are available: vehicle ID, model, battery power expressed in kW, the battery capacity expressed in kWh, the connector type, and the state of charge of battery expressed in percentages. The following types of EVs are considered: (a) Renault ZOE with a battery capacity of 22 kWh and a maximum charge power of 22 kW; (b) Renault ZOE with a battery capacity of 41 kWh and a maximum charge power of 22kW and (c) Nissan LEAF with a battery capacity of 24 kWh and a maximum charge power of 7 kW.

*We started the evaluation process by first setting up the configuration environment* (see Table 1). In our study, we considered two scenarios: one for charging electric vehicles when the microgrid has excess energy and the other for discharging electrical vehicles when there is a shortage of energy in the microgrid. In both scenarios, we considered 6 charging stations, 30 electric vehicles and a scheduling interval of 5 time slots.

**Table 1.** Environment configuration and scheduling operation type.

| Model id | Environment configuration | Scheduling Operation Type |
|---|---|---|
| L1 | # Charging stations = 6, # EVs = 30, DR | Charge |
| L2 | program period T=5 / t=1 | Discharge |

We then generated the curves for energy demand and for renewable energy production that need to be balanced through electric vehicle (EV) scheduling. The two curves were generated by simulation.

*Next, we moved on to Deep Q-Network (DQN) training.* In the training process we considered the following input features for model learning: the number of available charging stations, the number of EVs waiting for charging/discharging, the time slots for scheduling, the state of charge (SoC) of each EV, the capacity of each EV, and the energy available in each time slot representing the energy available for scheduling during that specific hour. This energy value dynamically updates every hour, by: (i) subtracting the energy needed to charge electric vehicles allocated to charging stations at that hour from the initial target energy value provided by the network operator in the case of charging operation or (ii) adding the energy discharged from electric vehicles allocated to charging stations at that hour to the initial target energy value provided by the network operator in the case of discharging operation. In the training process, the DQN relies on the replay memory which stores the agent's past experiences. After completing each episode, the agent selects a batch of experiences from this memory and uses it to train the network. The training process of Deep Q-Network (DQN) involves adjusting several hyperparameters in the learning process, namely memory size, learning rate, epsilon decay and batch size. To identify the optimal values of these hyperparameters we used a trial-and-error approach that involves testing different values for these parameters, observing their impact on model performance, and selecting the values that produced the best results. Table 2 shows the configurations tested and the results obtained in terms of minimum loss and maximum reward for 160000 episodes for each considered configuration.

**Table 2.** Configuration of network hyperparameters and obtained results

| Hyperparameters Tunning | | | | Results | |
|---|---|---|---|---|---|
| Memory Size | Batch size | $\varepsilon_{decay}$ | $\eta$ | Loss | Reward |
| 50000 | 1000 | 0.99996 | 0.001 | 43860637 | -107799 |
| 50000 | 700000 | 0.9995 | 0.001 | 4693810 | -106921 |
| 500000 | 15000 | 0.99996 | 0.001 | 49102022 | -102242 |
| 50000 | 700000 | 0.99996 | 0.005 | 88181132 | -104005 |
| 50000 | 700000 | 0.99996 | 0.01 | 252101836 | -103339 |
| 50000 | 700000 | 0.9997 | 0.001 | 82704 | -3852 |
| 600000 | 30000 | 0.99996 | 0.001 | 78076 | -4050 |
| 700000 | 50000 | 0.99996 | 0.001 | 14587 | 654 |

In the case of memory size, we started with a small memory size, which we gradually increased and looked at how it influenced the agent's learning capacity. In the case of a small size, the number of experiences the agent can store is limited, and as a result, due to insufficient space exploration, the policy learned is suboptimal. In the case of a large memory size, the agent accumulates a wide range of experiences, which leads to an improvement in exploration and a reduction in the occurrence of the phenomenon of overfitting. However, too large values lead to increased computational costs. Based on these findings and considering the dimension of the environment in which the agent learns, we've settled the value for memory size of 700,000.

The epsilon decay value was chosen as 0.99996 to promote exploration at the beginning of training and progressively reduce random decisions as the model learns. Using this value, the agent does more exploration in the initial episodes, thus gathering a wide range of transition experiences. As more episodes pass, epsilon decay gradually decreases, meaning that over time, the agent takes fewer random actions and begins to make decisions based on information learned from the neural network. The learning rate (η) value of 0.001 was chosen because smaller updates to model weights lead to more stable learning in complex environments, even though more training episodes are required to converge to optimal policy.

The batch size value was chosen equal to 50000 as it gives the best results for loss and reward metrics. A higher value implies a learning process that converges more slowly (i.e. longer training time) and also does not effectively improve the accuracy of the results, while a smaller value would imply a learning process that converges faster but gives results less accurately.

After establishing the optimal configuration for network hyperparameters *we proceeded to evaluate the quality of the training model*. To validate the training model, we have used two metrics, namely the average rewards and the loss function. Figure 3 shows the evolution of average rewards during L1 and L2 model training for charging and discharging operations. In the case of L1 model for charging we can see that the average rewards of the learning model stabilize around 140000 episodes, then at 150000 there is a temporary worsening of the received rewards, followed by a return and stabilization of the rewards starting from 16000 episodes. This situation may be due to the complex environment in which the agent operates, which may face situations in which its performance temporarily degrades until it adapts to the new conditions. In the case of the L2 model, a significant increase in rewards is observed between 40 000 and 100 000 episodes, followed by a smaller increase between 100 000 and 120 000 episodes. Starting from episode 120 000, the reward stabilizes, registering only a few small fluctuations. Like the L1 model, these small fluctuations are due to the complexity of the environment in which the agent operates and exploration vs. exploitation trade-off.

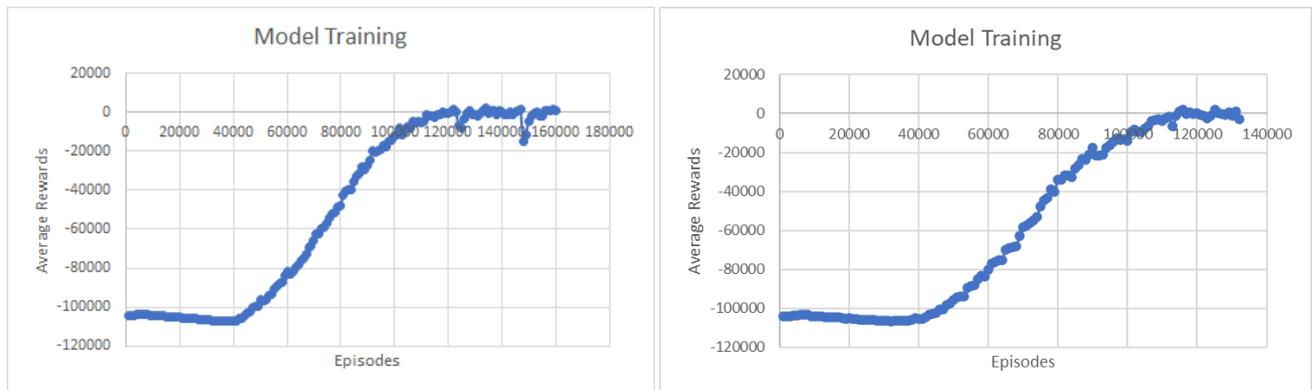

**Figure 3.** Average rewards for L1 and L2 models training.

Looking at the loss evolution presented in Figure 4, we observe a common trend followed by both the L1 and L2 models. At first the loss decreases as the models learn, and then after many episodes it starts to stabilize. More precisely, in the case of the L1 model, the loss starts to stabilize around 57,000 episodes, and in the case of the L2 model, around 37,000 episodes. The need for many episodes to stabilize loss derives both from the complexity of the environment in which the agent operates and from the trade-off of exploration versus exploitation. The agent needs more episodes to accumulate learning experiences and strike the right balance between exploration and exploitation.

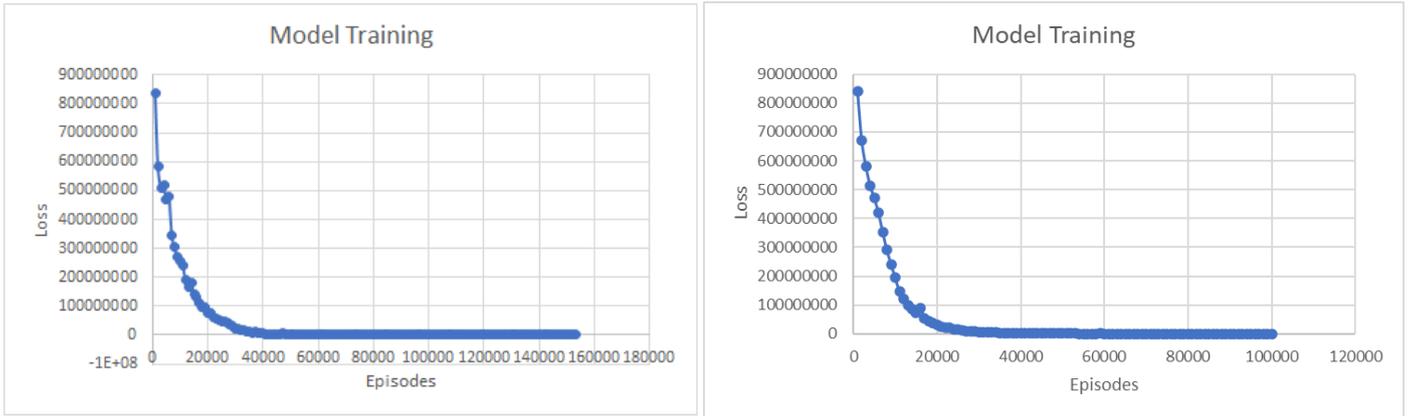

**Figure 4.** Average loss for L1 and L2 models.

*Finally, during DQN testing, we assessed the quality of the learned policy.* For this, we examined the percentage of allocations of electric vehicles at charging stations with maximum value of the reward, in relation to the total number of EVs allocations per time interval and epochs variations. We have also, examine the similarity between the curve derived from scheduling of electric vehicles at charging stations and the target energy curve provided by the grid operator using the Pearson correlation coefficient:

$$Pearson\big(E_{MG}(T), E_{Schedule}(T)\big) = \frac{\sum_{i=1}^{n}(E_{target}(t_i) - \overline{E_{target}})(E_{EV}(t_i) - \overline{E_{EV}})}{\sqrt{\sum_{i=1}^{n}(E_{target}(t_i) - \overline{E_{target}})^2} * \sqrt{\sum_{i=1}^{n}(E_{EV}(t_i) - \overline{E_{EV}})^2}} \quad (17)$$

In formula 17, $n$ is the number of energy values corresponding to the time interval during which the EVs' scheduling is performed, $E_{target}(t_i)$ and $E_{EV}(t_i)$ are the energy data points in the goal energy curve and the curve resulted by scheduling EV at charging station, $\overline{E_{target}}$ and $\overline{E_{SV}}$ are energy sample means computed as:

$$\overline{E_{target}} = \frac{1}{n}\sum_{i=1}^{n}(E_{target}(t_i)) \quad (18)$$

$$\overline{E_{EV}} = \frac{1}{n}\sum_{i=1}^{n}(E_{EV}(t_i)) \quad (19)$$

Figure 5 presents the percentage of optimal allocations of EVs versus sub-optimal ones for the two learned policies corresponding to the charging and discharging scenarios. In the case of the charging scenario, we observe that in 70.8% of cases, all vehicles are optimally allocated (that is, the trained model optimally manages EVs), in 17.5% of cases 1 EV is suboptimal allocated, and in 5.6% of the cases, two EVs are suboptimal allocated. Only a small percentage of cases features more EVs suboptimal allocations (i.e., 6.1%). This means that the learned model makes good decisions, since it can optimally allocate EVs in most cases. Even when suboptimal allocations occur, their frequency is relatively low, indicating that the model has learned a policy that works well for this scenario. In the case of L2 model, in 82.9% of cases EVs are optimally allocated to charging stations, in 14.4% of cases there is only one suboptimal allocation, and in the case of 2.5% there are two suboptimal allocations. As in the case of the L1 scenario, the policy learned in the case of the L2 scenario can make optimal decisions in most cases regarding the allocation of electric vehicles to the charging stations.

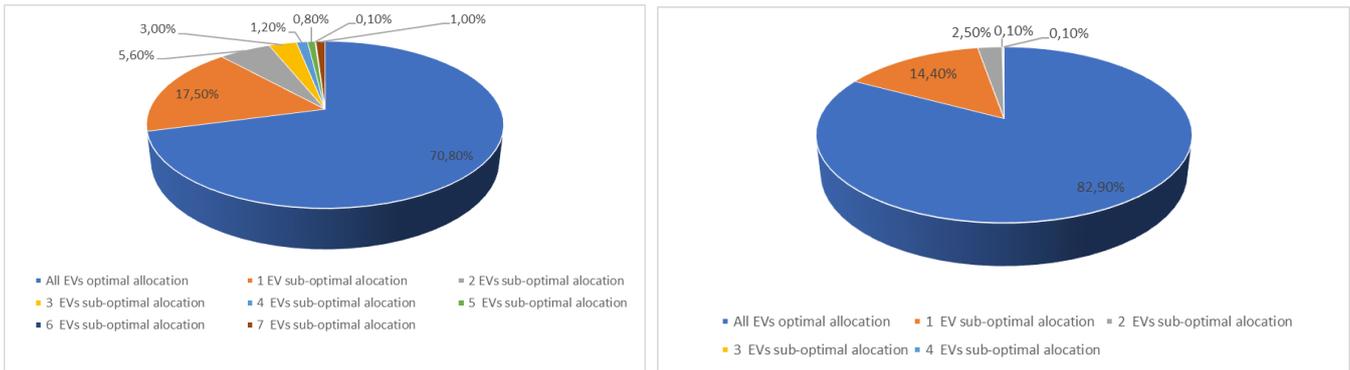

**Figure 5.** The percentage of optimal / sub-optimal allocations of EVs for the two policies.

The tracking of the evolution of the percentage of optimal allocations of electric vehicles over episodes is shown in Figure 6.

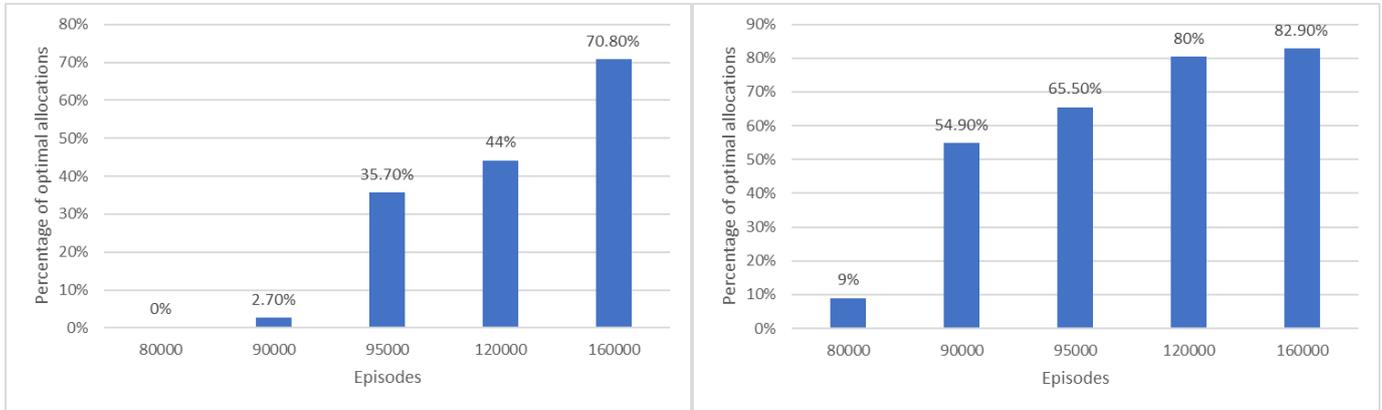

**Figure 6.** The evolution of the percentage of optimal allocations of electric vehicles over episodes for L1 and L2 model.

Examining the results obtained in the case of the L1 model, we noticed that after 9,500 episodes, the algorithm can optimally allocate EVs in 35.7% of cases, reaching an optimal allocation of 70.8% after 160,000 episodes. Similarly, for the L2 model, 82.9% optimal EVs allocation begins after 160,000 episodes. Consequently, in both models, the percentage of optimal allocations for EVs increases with each passing episode throughout the learning process, which demonstrates the effectiveness of the learned policies.

The final step of learned policy evaluation involved assessing the effectiveness of EV allocations at charging stations in each time slot throughout the scheduling period. Figure 7 presents the curve resulted by scheduling EVs at charging stations together with the target energy curve provided by the grid operator as well as the baseline curve in the case of charge and discharge scenarios. Analyzing these graphs, we noticed that for both learned models (i.e., L1 and L2), the curve obtained from scheduling EVs closely follows the target energy curve provided by the grid operator. Also, when we schedule the charging/discharging of electric vehicles with our method, the resulting energy curve is much closer to the target energy curve. EV scheduling for charging/discharging significantly improves charging/discharging capacity compared to the baseline scenario where EV charging/discharging is done without any prior appointment.

In the charging scenario, the variation between the energy required to charge electric vehicles and the energy supplied by the network operator in each time slot does not exceed a maximum deviation of 6 kW, indicating an optimal allocation of electric vehicles at charging stations. This high level of alignment between the two curves is reflected by the value of the Pearson correlation coefficient of 0,9926, which indicates a strong similarity between the energy curves. In the discharge scenario, the situation is similar. In this case, the Pearson correlation coefficient shall be 0,9937 and the variation between the energy discharged by electric vehicles and the network operator's energy requirements during each time slot does not exceed a maximum deviation of 4 kW.

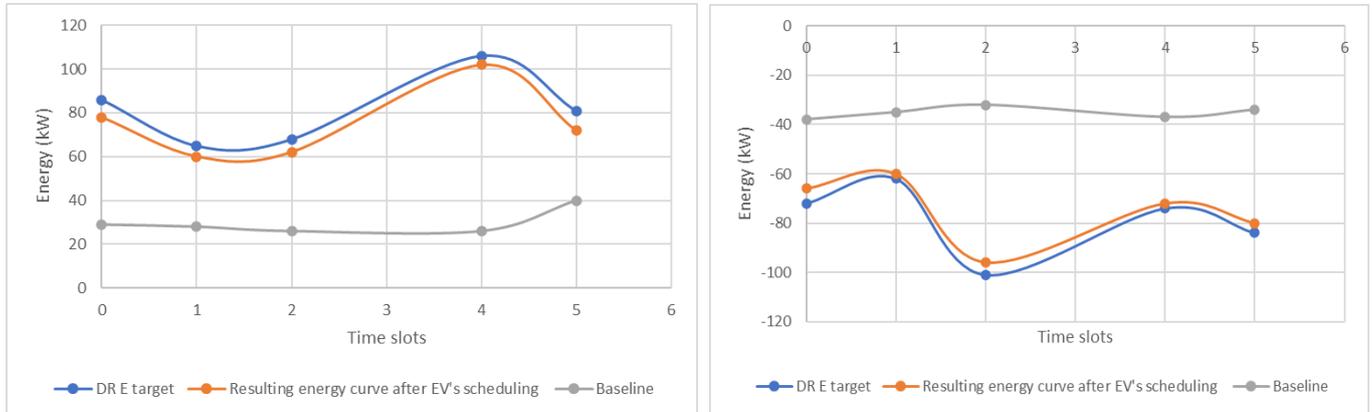

**Figure 7.** Comparison between the microgrid energy curve and the resulting curve after scheduling the EVs in the case of M5 model.

## 5. Conclusions

In this paper, we have proposed a Deep Q-Learning solution for enabling EVs to coordinate and participate in DR programs by scheduling the EVs charge and discharge actions to meet a target energy profile. We have adapted the Bellman Equation to evaluate the EVs scheduling state and defined a Q-function to determine the action's effectiveness in terms of rewards. We have represented the Q-function using a neural network to learn from for dynamic EVs scheduling scenarios by successively training the available actions and using Epsilon-greedy algorithm balances exploitation and exploration of the state space.

Our proposed solution offers a promising approach to EVs participation in DR programs, which can result in significant benefits for both the distribution system operator and EV owners. The results show that the EVs are scheduled to charge and discharge effectively their collective energy profile following accurately with a Person coefficient of 0.99 the energy profile provided as a target in DR. Moreover, the rewards and loss converge to good values, showing the effectiveness of the quality of the learning policy defined for EVs scheduling as well as of the Q-function in improving the agent's decision-making in the defined microgrid environment. Finally, a limitation of our approach is the fact that the learning process needs a substantial number of episodes to achieve loss and rewards because of the complexity of the environment. There are many states to navigate to gather learning experience due to scheduling combinations of EVs, charging stations, and trade-offs between exploration and exploitation.

As future work, we intend to investigate new EVs scheduling solutions by considering iteratively each time slot in the DR program window and not the entire interval as in the current approach. This type of solution may reduce the number of episodes needed for loss and reward convergence while improving the agent responsiveness to changes in the defined EVs scheduling environment and promoting greater dynamism in decision-making. In addition, we plan to investigate alternative Deep Q-learning network architectures to allow the agent to learn and adapt more efficiently with fewer episodes.

**Acknowledgments:** This work has been conducted within the DEDALUS project grant number 101103998 funded by the European Commission as part of the Horizon Europe Framework Programme.